# Benchmarking and Domain Adaptation of Automatic Speech Recognition (ASR) for Adolescent Health Communication in Ghanaian Languages

Stephen E. Moore[*,1,2], Akwasi Asare[2], Mich-Seth Owusu[2], Paul Azunre[2], Joel Budu[2], Lawrence A. Adu-Gyamfi[2]

[1]Department of Mathematics, University of Cape Coast, Ghana

[2]Ghana Natural Language Processing, Cape Coast, Ghana

[*]Corresponding Author: stephen.moore@ucc.edu.gh

## ABSTRACT

This paper presents an end-to-end study of automatic speech recognition (ASR) for adolescent health communication in three Ghanaian languages (Twi, Dagbani, and Ewe). The work proceeds in three connected stages; First, we benchmark five ASR systems (three language-specific Wav2Vec2 models and two multimodal LLMs, Gemma 3n and Gemma 4) on a general-domain Bible corpus and a Youth Adolescent Sexual and Reproductive Health (ASRH) Domain ASR dataset, using Character and Word Error Rate (CER, WER). Second, guided by the benchmark, we perform supervised domain adaptation: although Gemma 4 was the strongest zero-shot candidate, fine-tuning it proved computationally infeasible, so we pivoted to the compact Qwen3-ASR-0.6B, fine-tuned on a large Ghana Bible corpus (~90k samples) and evaluated strictly on held-out human-collected in-domain audio. Fine-tuning reduced WER on every language, most dramatically for Ewe (WER from 109.3% to 64.8%, a drop of 44.5 pp; CER from 65.1% to 24.9%). Third, we validate the work through KasaHealth, a live voice-first ASRH application deployed in all three languages, complemented by Senti-Check, a technical evaluation harness. KasaHealth was tested by 50 community respondents and achieved a 100% chat-approval rate, a 72% Good-or-Excellent translation rating, and a 92% would-recommend rate, while surfacing the domain gaps that most constrain real-world use. Across all three stages the evidence converges: for these languages the binding constraint is validated in-domain data, not model capability or computation.

## 1. INTRODUCTION

Access to health information in one's native language is foundational to meaningful participation in health services, particularly for adolescents in multilingual communities [1]. In Ghana, where over 80 languages are spoken, digital services have historically targeted English-speaking, literate populations, leaving millions of indigenous-language speakers without adequate access [2,3]. Three languages anchor this study: Twi, an Akan dialect cluster and the most widely spoken first language and national lingua franca; Dagbani, the principal language of the Northern Region, spoken by several million people; and Ewe, spoken across the Volta Region and into Togo and Benin [4–6]. Despite their importance, these languages remain severely underrepresented in NLP and commercial speech technology, so voice interfaces, especially valuable where literacy is lower or stigma limits text-based communication, are not reliably available to their speakers. ASRH is an especially sensitive setting: accurate transcription of health terminology directly affects whether young people and health workers can trust a voice interface.

This paper reports a complete arc from measurement to deployment. We benchmark available models across a general domain (Bible speech) and the target ASRH domain; use that benchmark to select and execute a domain-adaptation strategy, documenting a pivot from a multimodal LLM to a compact deployable ASR model; and validate the result with real users through the KasaHealth application, characterising the remaining gaps. All datasets, model weights, and tooling are released open-source. Our contributions are: (i) a five-model, two-domain ASR benchmark for Twi, Dagbani, and Ewe introducing Gemma 4 as a newly competitive zero-shot baseline; (ii) a documented adaptation pathway from an infeasible Gemma fine-tune to a tractable Qwen3-ASR-0.6B fine-tune on modest (T4) hardware with strict train/evaluation domain separation; (iii) a user-validated deployment (KasaHealth) plus a technical harness (Senti-Check); and (iv) a fully open-source, reusable pipeline that other West and Central African language communities can adopt.

## 2. BACKGROUND AND RELATED WORK

*ASR for low-resource languages:* Progress has historically been constrained by scarce annotated audio, complex tonal phonology and rich morphology, and the absence of standardised orthographies [7]; traditional systems needed hundreds to thousands of hours of transcribed speech [8,9]. Self-supervised learning changed this: wav2vec 2.0 [10] learns acoustic representations from unlabeled audio and can be fine-tuned on as little as ten minutes of labeled speech, and its multilingual XLSR variant [11] enables cross-lingual transfer. In health settings the bar is higher. Afonja et al. found that although general WER was often low, errors on medical entities were substantially higher, with domain fine-tuning improving medical WER by 25–34% relative [12], while Blocker et al. reported isiXhosa CER of 43–51% in primary care even where English approached human transcription [13]. General benchmarks are therefore insufficient proxies for health-domain performance. *Ghanaian-language NLP and zero-shot LLMs*: GhanaNLP has produced publicly available Wav2Vec2 ASR models and datasets for Twi, Dagbani, and Ewe [2,4,5,14–16] exposed through the Khaya AI initiative. Multimodal LLMs offer zero-shot transcription without task-specific training: Gemma 3n performed poorly across all three languages [17], whereas Gemma 4 (evaluated here for the first time) is substantially upgraded. As the adaptation results below show, however, competitiveness as a zero-shot baseline does not imply tractability as a fine-tuning target under realistic compute budgets.

*The ASRH domain gap:* Domain mismatch raises error rates through unfamiliar vocabulary, speaking styles, and prosody [18]. ASRH is a particularly demanding domain: reproductive anatomy, contraception, STIs, and adolescent development require precise terminology rarely present in Bible-derived corpora, and local-language discussion involves socially constructed euphemisms that diverge from formal medical vocabulary. This structural gap motivates the adaptation and deployment stages that follow.

## 3. METHODOLOGY

### 3.1 Datasets

Two benchmarking domains were used, both public on the Hugging Face Hub under the Ghana NLP organization: Bible datasets [14–16] and Youth ASRH Domain ASR datasets [19–21]. The Youth ASRH

ASR datasets (ghananlpcommunity/UNICEF-Ghana-{Twi,Dagbani,Ewe}-ASR) pair audio with transcriptions drawn from ASRH-relevant materials; the Bible datasets (asante-twi-bible-speech-text, dagbani- and ewe-bible-audio-text-tts) provide broader general-domain vocabulary and a reference domain closer to the training distribution. A synthetic ASRH-domain text dataset grounded in UNICEF Ghana's adolescent counselling manual, covering ASRH, Child Protection, Education, and Environmental Concerns in English and all three languages, was also generated and released [22]. Benchmarking uses 50-sample subsets per dataset per language, up from a preliminary 30-sample evaluation, improving statistical reliability.

### 3.2 Models and Benchmarking Methodology

Five models were benchmarked. dagbani_wav2vec2 and ewe_wav2vec2 are Wav2Vec2-BERT (w2v-bert-2.0) models fine-tuned on their languages [10,23,24] ; twi_w2v_bert adds a BERT language-model decoding component for improved word-level accuracy [23,25]. gemma-3n-E2B-it [17] and gemma-4-E2B-it [26] are instruction-tuned multimodal LLMs evaluated fully zero-shot. CER and WER [27–29] were computed with jiwer as the mean of per-sample scores; a WER above 100% indicates hallucination or poorly calibrated decoding. Audio was resampled to 16 kHz; references and hypotheses were lowercased, and whitespace stripped. Wav2Vec2-BERT models used greedy CTC decoding (no beam search, no external LM, batch 16); Gemma models ran via Unsloth FastModel in bfloat16 with greedy decoding and up to 200 new tokens, using a minimal transcription prompt with no language specification or few-shot examples. Wav2Vec2 models were evaluated on native and both non-native languages; Gemma on all six conditions.

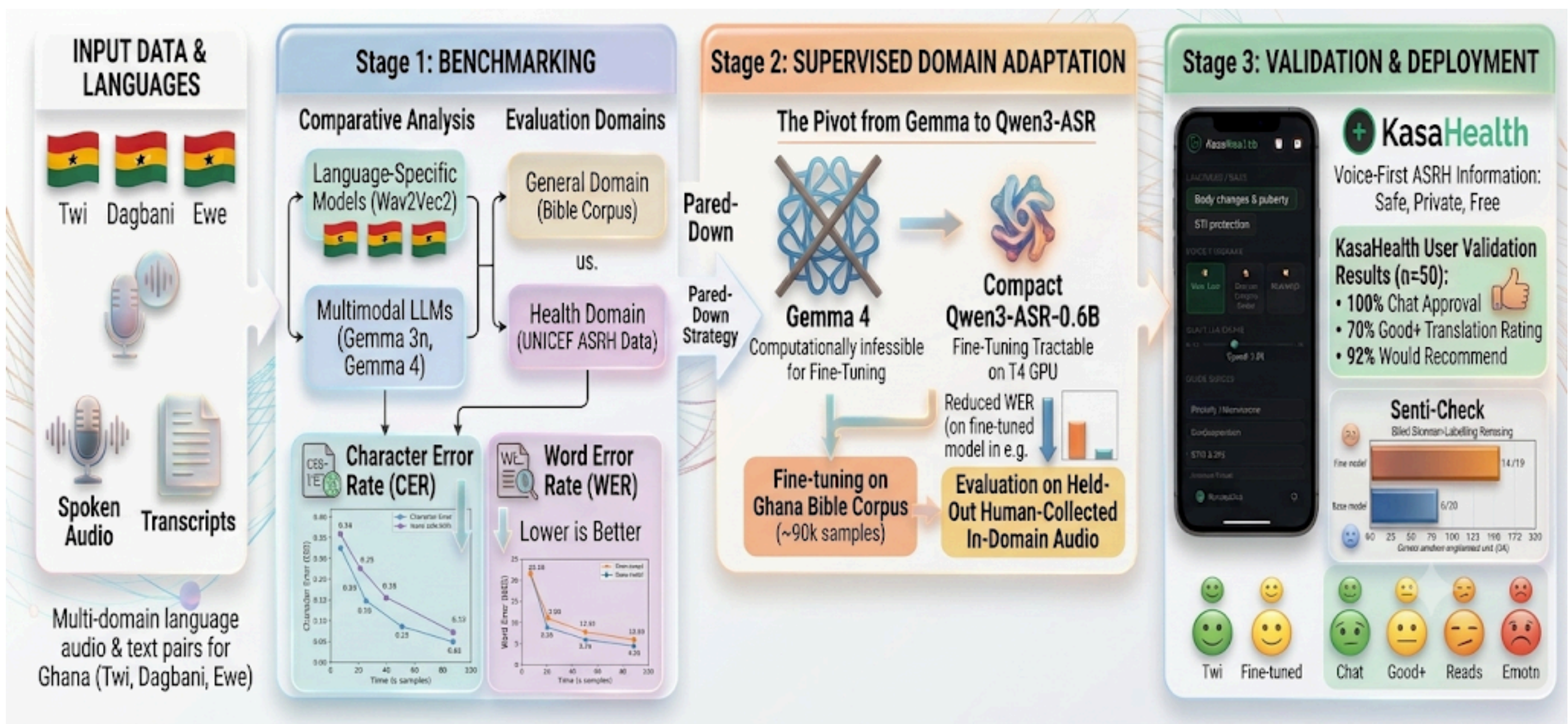

### 3.3 Audio Collection and the Pivot from Gemma to Qwen3-ASR

A Python offline GUI recorder [30], designed for low-connectivity settings, collected transcribed audio from community participants: ~30 minutes per language in Phase 1 (the 50-sample benchmarking set) and up to two hours per language in Phase 2, reserved strictly as the held-out evaluation set. Gemma 4 E2B was initially selected as the fine-tuning target based on its cross-dataset stability (Section 4.1), but its multimodal architecture and parameter count exceeded the available GPU resources. The work therefore adopted Qwen3-ASR-0.6B instead, a compact multilingual model whose smaller footprint made supervised fine-tuning tractable on T4 hardware and whose language-prefix output format enables simultaneous transcription and language identification. Training data and evaluation data were deliberately separated: rather than mixing synthetic TTS audio with human recordings, the model was trained on a large out-of-domain Ghana Bible corpus, with the human-collected Phase-2 audio reserved strictly for evaluation, so that reported metrics reflect generalisation to the adolescent domain rather than training overlap. Table 1 summarises the fine-tuning configuration; training loss fell from approximately 150 to below 1.0 and evaluation loss from 0.32 to 0.09, indicating convergence without overfitting.

Table 1. Qwen3-ASR-0.6B fine-tuning configuration.

| Parameter | Value |
|---|---|
| Training dataset | ghananlpcommunity/ghana-bible-combined-90k-twi-ewe-dagbani (~90k samples; 98/2 train/eval) |
| Base / released model | Qwen/Qwen3-ASR-0.6B → ghananlpcommunity/qwen3-asr-0.6b-ghana-twi-ewe-dagbani |
| Epochs / batch / LR | ~3.2 (to convergence) / effective 32 (4×16) / 2e-5, warmup + cosine decay |
| Final eval loss / hardware | ~0.09 / T4 GPU (Google Colab) |
| Evaluation set | Human-collected in-domain audio (up to 2 h/language, Phase 2) |

## 4. RESULTS AND DISCUSSION

This section reports the three stages of evaluation in turn: the cross-model benchmark (Section 4.1), the domain-adaptation fine-tuning (Section 4.2), and the user acceptance testing (Section 4.3). Section 4.4 synthesises the implications, and Section 4.5 lists the open-source resources released with the work.

### 4.1 Benchmarking

Table 2 reports CER and WER for all five models across both datasets, and Figure 1 and Figure 2 summarise the WER and CER results respectively.

Table 2. CER and WER for all benchmarked models (50 samples per dataset). Wav2Vec2 rows report native-language evaluation; Gemma rows are per language. Lower is better.

| Model | Lang | Dataset | CER % | WER % | Model | Lang | Dataset | CER% | WER% |
|---|---|---|---|---|---|---|---|---|---|
| dagbani_w2v | Dag | Bible | 25.2 | 74.6 | gemma-3n | Ewe | Bible | 199.1 | 215.6 |
| dagbani_w2v | Dag | Youth ASRH ASR | 19.5 | 56.4 | gemma-3n | Ewe | Youth ASRH ASR | 59.7 | 114.4 |
| ewe_w2v | Ewe | Bible | 17.9 | 60.8 | gemma-3n | Twi | Bible | 326.2 | 337.4 |
| ewe_w2v | Ewe | Youth ASRH ASR | 18.9 | 64.0 | gemma-3n | Twi | Youth ASRH ASR | 75.6 | 132.3 |
| twi_w2v_bert | Twi | Bible | 17.0 | 53.6 | gemma-4 | Dag | Bible | 51.2 | 98.4 |
| twi_w2v_bert | Twi | Youth ASRH ASR | 20.9 | 62.2 | gemma-4 | Dag | Youth ASRH ASR | 55.4 | 96.8 |
| gemma-3n | Dag | Bible | 240.3 | 344.0 | gemma-4 | Ewe | Bible | 47.2 | 98.8 |

| gemma-3n | Dag | Youth ASRH ASR | 93.0 | 141.1 | gemma -4 | Ewe | Youth ASRH ASR | 50.1 | 98.3 |
|---|---|---|---|---|---|---|---|---|---|
| | | | | | gemma -4 | Twi | B/ASR H | 49.2/44. 3 | 92.5/92. 6 |

*Figure 1. WER by model: Bible vs Youth ASRH ASR.*

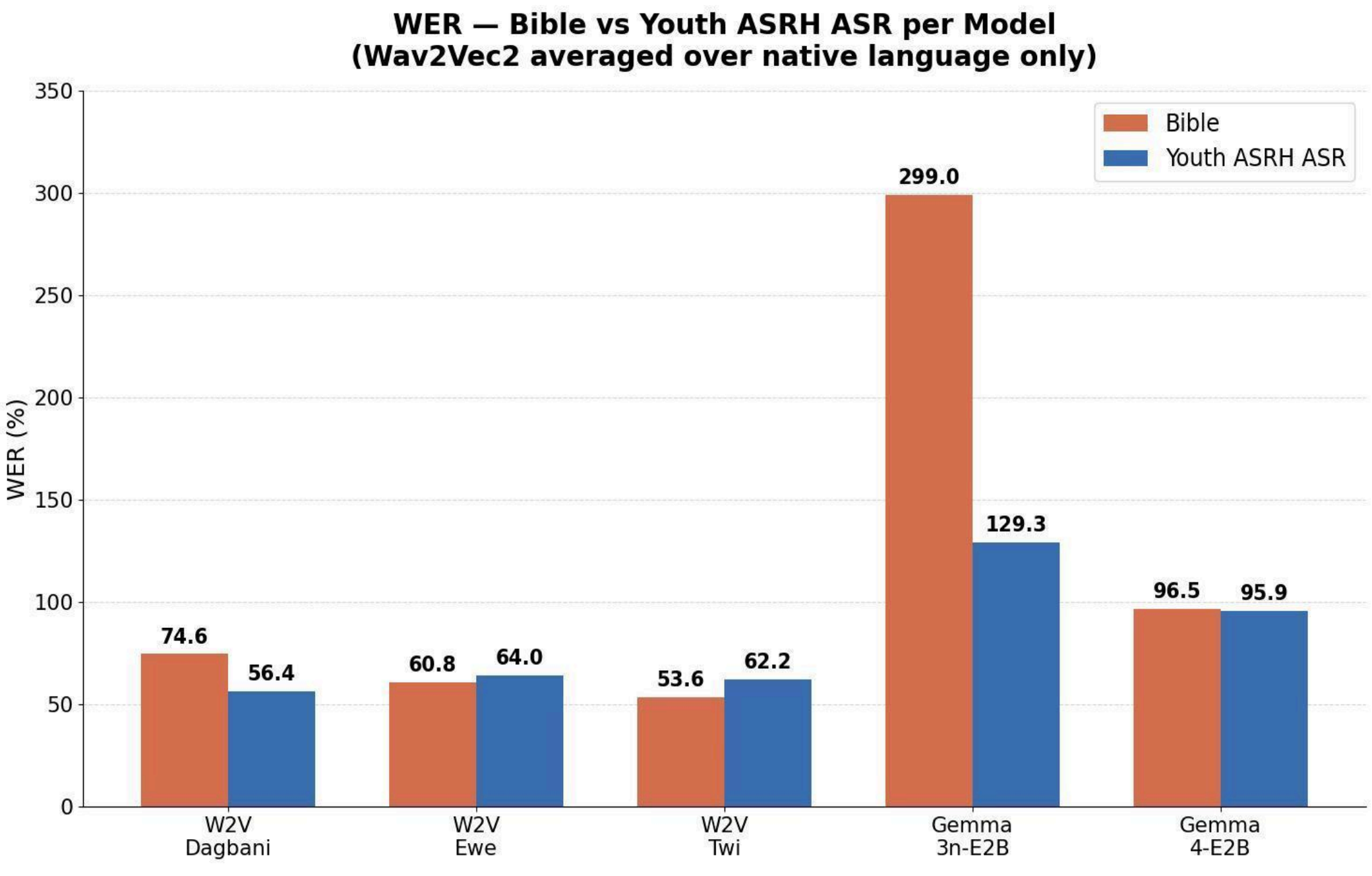

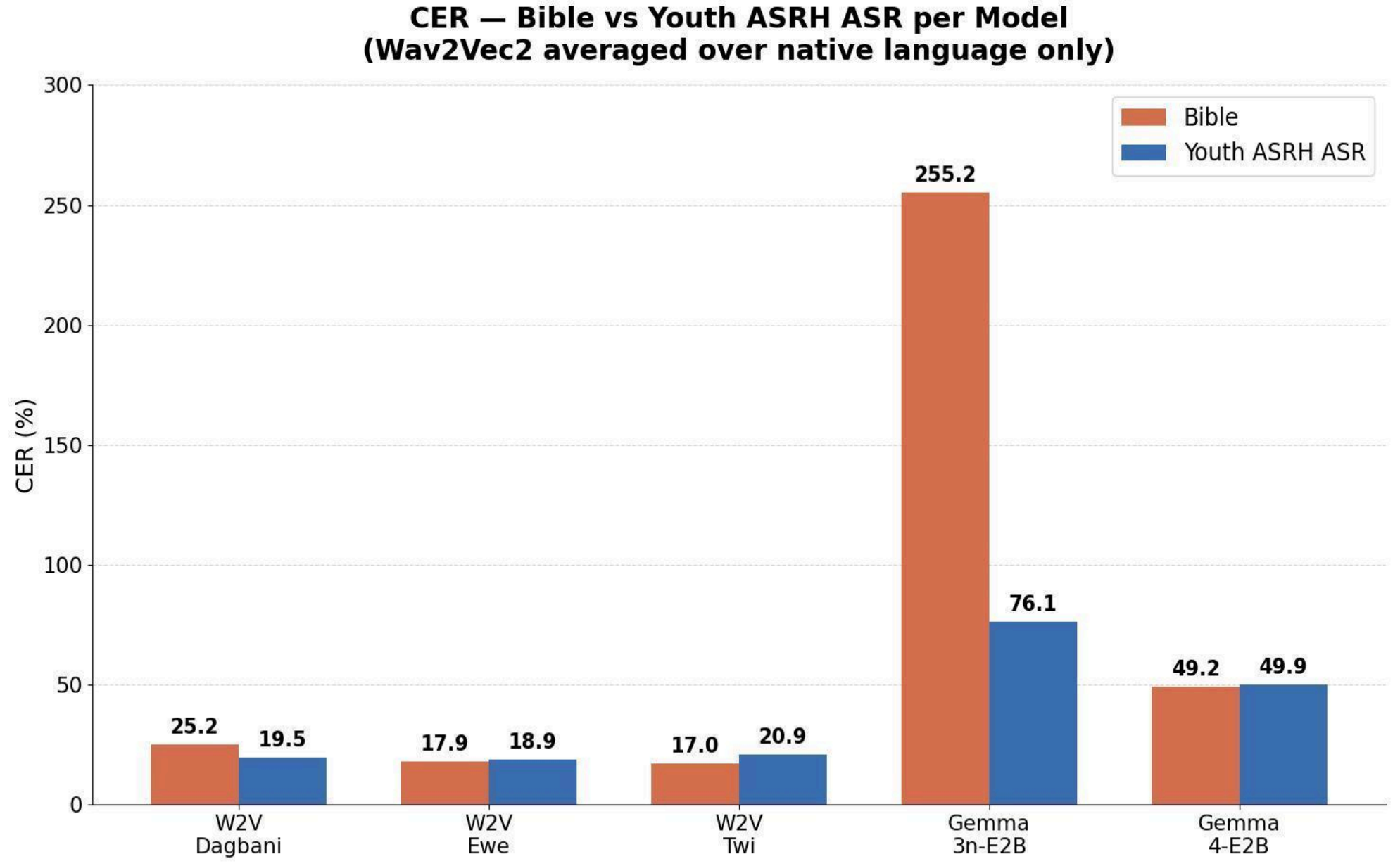

*Figure 2. CER by model: Bible vs Youth ASRH ASR. (Wav2Vec2 values averaged over native conditions; Gemma across all three languages.)*

#### *4.1.1 Model-by-Model Analysis*

*Wav2Vec2 language-specific models:* The three fine-tuned models perform strongly on their native languages. twi_w2v_bert records the best CER overall (17.0% Bible, 20.9% Youth ASRH); ewe_wav2vec2 achieves 17.9%/18.9% CER, a notably small ~1 pp domain gap although its WER gap is larger at 3.2 pp; and dagbani_wav2vec2 records 25.2%/19.5% CER, with the Youth ASRH set producing a lower CER than Bible, a reversal suggesting the Dagbani model is better calibrated to the acoustic characteristics of the health recordings. On non-native languages all three degrade sharply (CER 45–65%, WER consistently >95%), confirming that fine-tuning confers language-specific rather than broadly transferable capability.

*Gemma 3n: persistent limitations*: Gemma 3n exhibits very high error rates across all conditions, with CER on Bible from 199.1% (Ewe) to 326.2% (Twi) and WER from 215.6% (Ewe) to 344.0% (Dagbani), systematic hallucination in which the model generates text unrelated to the reference. Performance is markedly better on Youth ASRH data (CER 59.7% Ewe, 75.6% Twi, 93.0% Dagbani), consistent with the preliminary 30-sample results and likely reflecting the constrained vocabulary and topic distribution of the health recordings relative to the diverse register of biblical text [17], [25]. Given the extreme absolute error rates and the hallucination observed on Bible data, these lower values do not indicate genuine ASR competence.

*Gemma 4: a competitive zero-shot baseline*: Gemma 4 is the most notable result of the benchmark. Across all six conditions it achieves CER 44.3–55.4% and WER 92.5–98.8%, substantially and consistently better than Gemma 3n (a CER reduction exceeding 100 pp on Bible and 15–35 pp on Youth ASRH). Its WER is directly competitive with the fine-tuned Wav2Vec2 models (which reach 53.6–74.6% on Bible and 56.4–64.0% on Youth ASRH) despite receiving no language-specific training; the Wav2Vec2 models retain a clear CER advantage, but the WER convergence indicates Gemma 4 produces more coherent word-level output than its CER alone would imply. A notable feature is its stability: consistent CER and WER across languages and datasets with a domain gap of at most a few percentage points, reflecting multilingual pretraining. This cross-language and cross-domain stability motivated its initial selection as the fine-tuning target. Figure 3 presents the full CER matrix across all model and dataset combinations, including cross-lingual conditions.

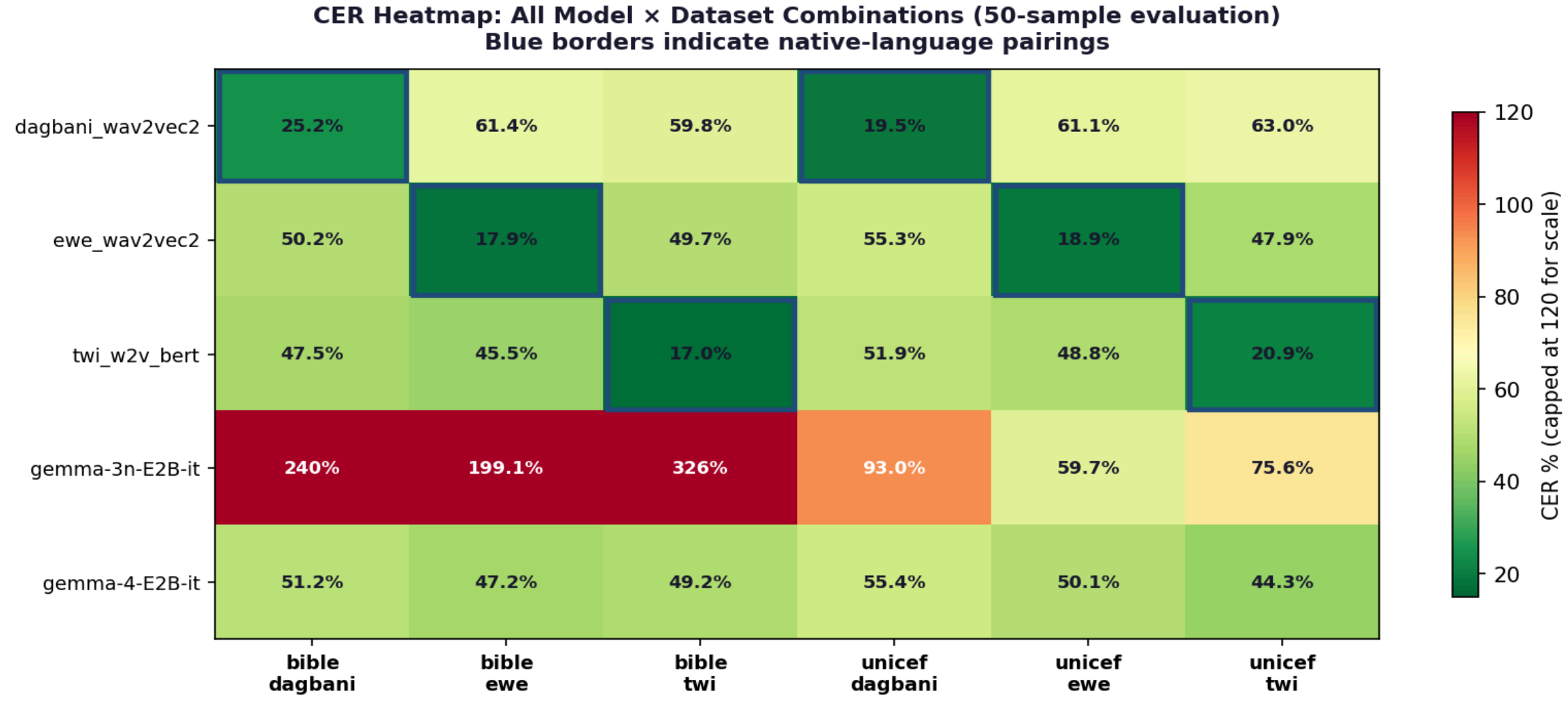

*Figure 3. CER heatmap across all model–dataset combinations. Blue-bordered cells are native pairings; Gemma 3n values >120% are capped for legibility.*

*The domain gap as a structural challenge*: Degradation on Youth ASRH data relative to Bible data is inherent to the current state of low-resource ASR here. The Bible corpus is the largest transcribed source for these languages, so Wav2Vec2 models inherit its vocabulary and acoustic profile and encounter distribution shifts elsewhere. The gap extends even to Gemma 4, indicating general multimodal pretraining has not closed the distributional gap between religious speech and health communication, motivating the adaptation stage.

### 4.2 Domain Adaptation: Fine-Tuning

The fine-tuned Qwen3-ASR-0.6B model was evaluated against its base version on held-out human-collected Youth ASRH-domain audio. Because the Bible training corpus and the in-domain evaluation set are disjoint in domain, these results measure generalization rather than memorization. Figure 4 compares base and fine-tuned WER and CER, and Table 3 reports the corresponding values.

Qwen3-ASR-0.6B: Base vs Fine-tuned — Ghana Adolescent Health Domain

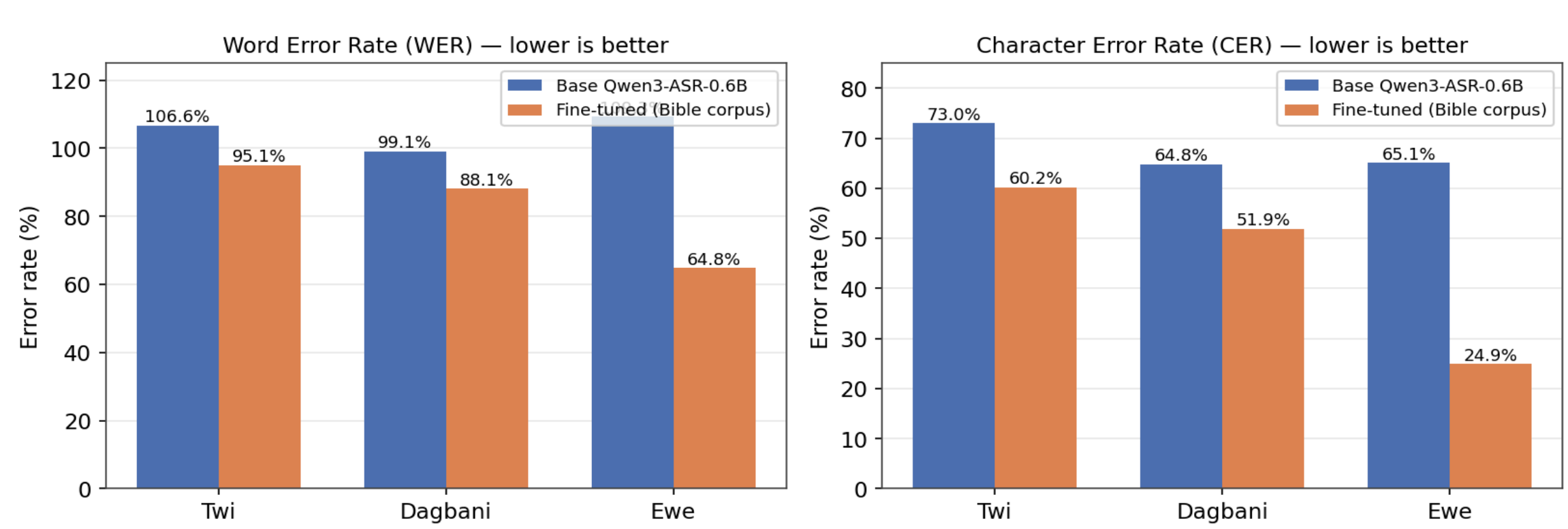

*Figure 4. WER (left) and CER (right) for the base and fine-tuned Qwen3-ASR-0.6B on held-out human-collected adolescent-domain audio.*

Table 3. Base and fine-tuned (FT) Qwen3-ASR-0.6B on held-out in-domain audio. A negative Δ indicates improvement.

| Language | WER Base | WER FT | WER Δ | CER Base | CER FT | CER Δ |
|---|---|---|---|---|---|---|
| Twi | 106.6% | 95.1% | −11.5 pp | 73.0% | 60.2% | −12.8 pp |
| Dagbani | 99.1% | 88.1% | −11.0 pp | 64.8% | 51.9% | −12.9 pp |
| Ewe | 109.3% | 64.8% | −44.5 pp | 65.1% | 24.9% | −40.2 pp |

Fine-tuning improves every language on both metrics. Ewe shows the most dramatic gain, with WER dropping 44.5 pp (from 109.3% to 64.8%) and CER dropping 40.2 pp (from 65.1% to 24.9%), while Twi and Dagbani show consistent ~11 pp WER improvements. The largest gains accrue to Ewe, the language with the weakest base performance, indicating that adaptation on out-of-domain data yields the greatest benefit for the languages furthest from the base model's training distribution. Residual error reflects both the inherent difficulty of low-resource ASR and the domain gap between the Bible training corpus and the adolescent-health evaluation set. Because the model improves despite training only on out-of-domain data, its in-domain performance is bounded chiefly by the quantity of in-domain material seen during training, which here was none. These results therefore represent a conservative lower bound, and a modest volume of in-domain training audio would be expected to improve them substantially.

### 4.3 User Acceptance Testing

Beyond error-rate improvement, a deployed system must be usable, trusted, and helpful to its intended users. Validation therefore combined two complementary applications. KasaHealth is a production, user-facing application evaluated by community respondents, and additionally serves as a reference for how a strong production pipeline performs in ASRH without domain-specific adaptation. Senti-Check is a technical harness that isolates the effect of the fine-tune. Because KasaHealth runs Ghana NLP's production Khaya pipeline rather than the fine-tuned model, the two are complementary rather than directly comparable.

#### *4.3.1 KasaHealth: Voice-First ASRH Application*

KasaHealth [31] is a voice-first health information application by GhanaNLP (Khaya AI model powered) delivering ASRH information in Twi, Dagbani, and Ewe through an ASR, LLM, and TTS pipeline built on production Khaya models. It most directly embodies the project's core objective of putting voice-based health information into young people's hands in their own language and is the flagship demonstration of the work. Before entry, every user was shown a disclaimer identifying the platform as an AI testing tool rather than a medical service, directing serious concerns to SHEplus Ghana, and confirming that sharing personal information was voluntary. Figure 5 shows the application interface.

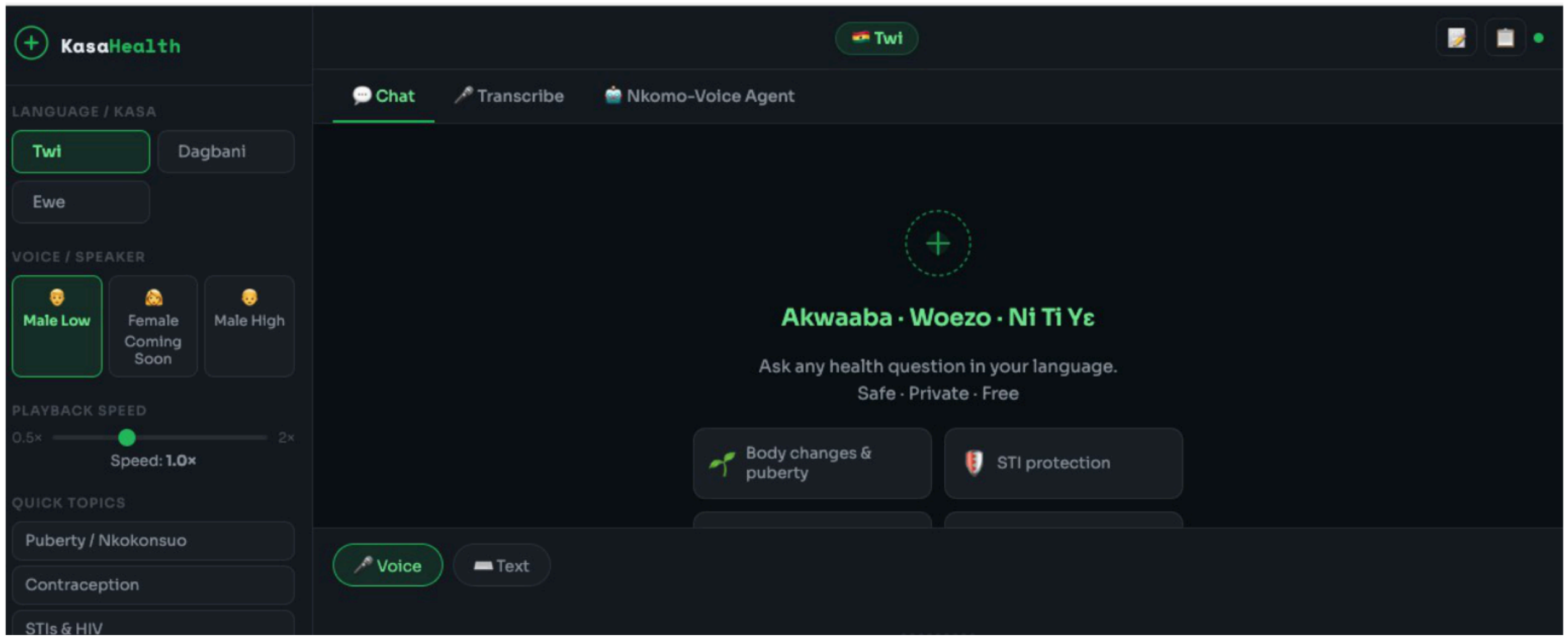

*Figure 5. The KasaHealth interface (Twi). Users select language and voice; the main panel offers voice/text input ("Hold to speak"), quick health topics (puberty, contraception, STIs & HIV, pregnancy, menstruation, consent & safety), and a chat with transcription, framed as "Safe · Private · Free".*

Testing ran 22–26 May 2026 with 50 respondents (24M, 26F), generating 107 items of correspondence (50 feedback-form submissions, 57 in-app ratings); 26 female respondents represent the primary intended user group. Of original testers, 92% would recommend the application and 75% rated answers Helpful or Very Helpful, with none rating them unhelpful. Figure 6 summarises component-level quality ratings across all 50 respondents.

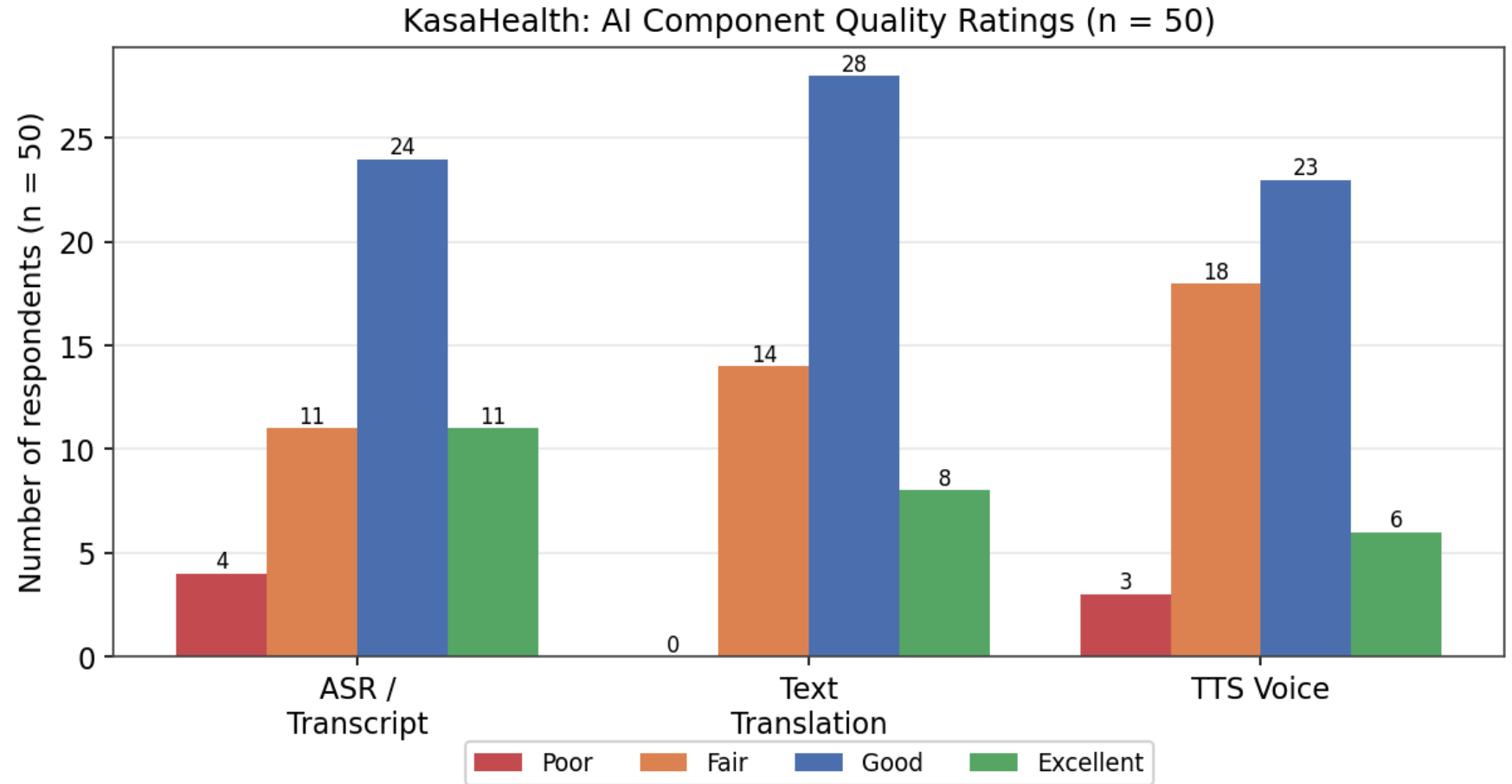

*Figure 6. KasaHealth AI component quality (n = 50). Translation received zero Poor ratings.*

a) ASR/transcript: 70% Good/Excellent (35/50). Translation: 72% Good/Excellent (36/50), zero Poor, the strongest component. TTS: 58% Good/Excellent (29/50), main issue Twi vowel-elongation artefacts.
b) Female respondents (n=26): 88% rated translation and 80% rated ASR Good/Excellent, the strongest of any subgroup, and encouraging for the primary audience.
c) In-app (57 interactions): all 26 health-question interactions received a thumbs-up (100% chat approval; Twi 11/11, Dagbani 14/14, Ewe 1/1). Of 29 flagged translation-review interactions, reason codes were dominated by bad translation (18, 62%), then incomplete answer (5), wrong language (2), other (2), wrong information (1). Ewe drew the most negative feedback (1 up/16 down), consistent with its residual benchmarking difficulty.
d) Key issues: Twi TTS vowel elongation ("anaa"→"anaaaaa"); a language-switching session-state bug; and ASRH vocabulary gaps where English health terms lack local equivalents, the signature of a domain gap, directly motivating targeted in-domain data collection.

#### *4.3.2 Senti-Check: Technical Validation*

Senti-Check [32] evaluates base and fine-tuned Qwen3-ASR-0.6B via a blind emotion-classification proxy: if a transcription preserves meaning, evaluators should identify the correct emotional tone from the transcript alone. Both models transcribe the same audio; transcripts receive one of 24 emotion labels; and six native-speaker testers (two per language, 20 samples each = 120 judgements/model) select the perceived emotion blind to model identity. As shown in Figure 7, the fine-tuned model doubled base accuracy (16/120, 13.3% vs 8/120, 6.7%). Absolute accuracy is low, as expected for a 24-class blind task, but the relative gain is consistent: strongest in Twi (from 3 to 9 of 40), with Dagbani improving (from 2 to 4 of 40) and Ewe stable (3 of 40), pending an upstream word-boundary fix. Table 4 contrasts the design and outcomes of the two applications.

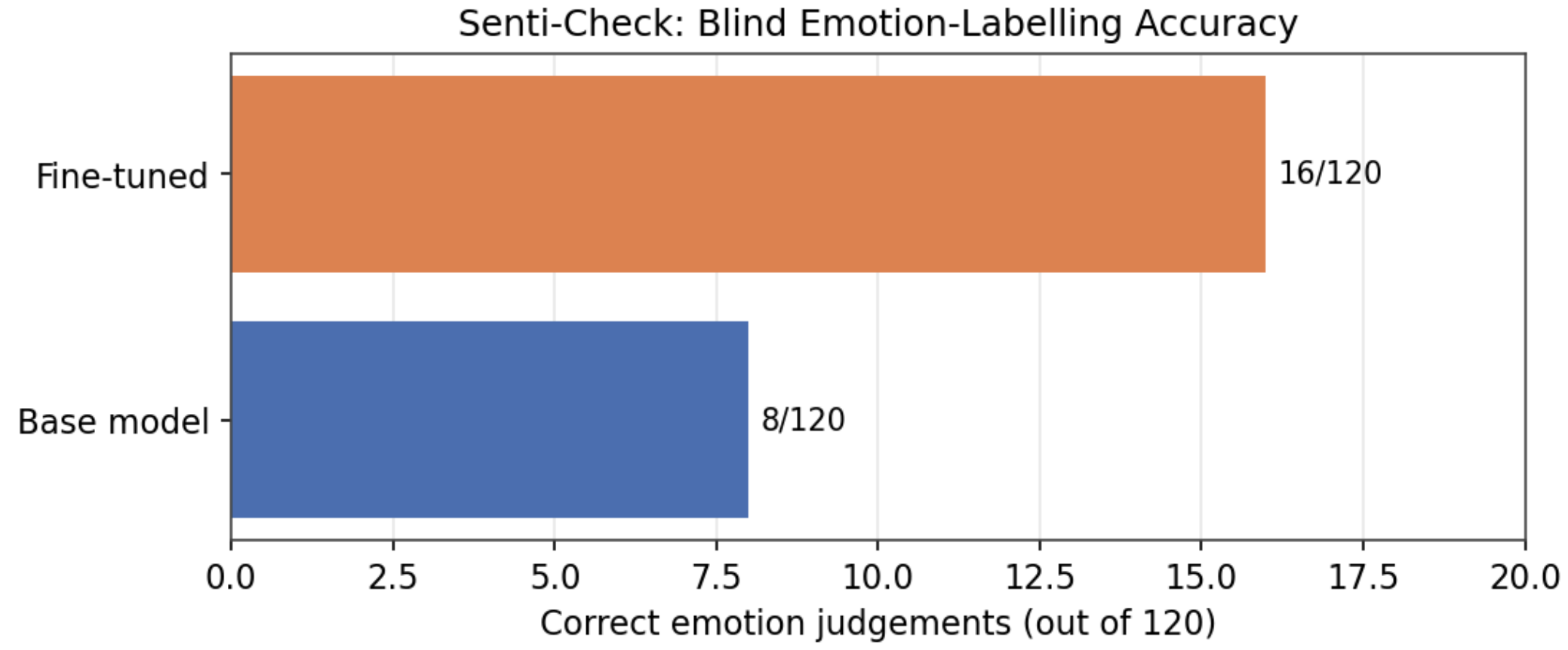

*Figure 7. Senti-Check blind emotion-labelling accuracy: fine-tuned 16/120 vs base 8/120 (2×).*

Table 4. KasaHealth and Senti-Check: two complementary evaluation applications.

| Dimension | KasaHealth (primary) | Senti-Check (validation) |
|---|---|---|
| Tests | Is the voice health app usable/helpful? | Does fine-tuning improve transcription quality? |
| Method / users | Survey + thumbs / 50 community (24M,26F) | Blind emotion (24 labels) / 6 testers |
| Models | Production Khaya pipeline | Base vs fine-tuned Qwen3-ASR-0.6B |
| Key result | 100% chat approval; 72% translation Good+ | Fine-tuned 2× base (16/120 vs 8/120) |

### 4.4 Discussion

The benchmarking, adaptation, and validation results converge on a common implication: the principal opportunity for improvement lies in data, followed by models, and only then compute. The limiting factor throughout was not model capability or compute but the availability of validated in-domain training data. The model was trained on a large but out-of-domain corpus of approximately 90,000 Bible samples because it was the largest high-quality transcribed resource available, and the recurring KasaHealth issues, namely ASRH vocabulary gaps and domain-specific translation errors, are characteristic of a domain gap rather than a fundamental model limitation.

*Data requirements:* Three kinds are most valuable. First, in-domain spoken audio for ASR: conversational health questions and counselling-style dialogue paired with human-verified transcriptions. Second, parallel health-terminology text for the translation and LLM components: English-to-local pairs covering ASRH terms, many of which have no standard local equivalent and which are the chief source of the translation errors observed in KasaHealth. Third, targeted natural-speech recordings for TTS, to address artefacts such as the Twi vowel elongation. In all cases the data should be speaker-diverse across gender, age, dialect, and recording conditions, mirroring the recorder methodology already validated here. As an indicative target, moving in-domain transcribed audio from the current ~2 hours per language (evaluation-only) toward the order of tens of hours per language available for training is a reasonable next milestone, though this is a direction, not a fixed prescription. Critically, well-validated in-domain data delivers gains out of proportion to its size relative to large out-of-domain corpora: a smaller, carefully validated in-domain set will usually outperform a larger, noisier one.

On models, KasaHealth shows that strong production-grade local-language models already deliver a usable, trusted end-to-end experience; the path is to domain-adapt efficient, deployable models (the Qwen3-ASR route) and apply targeted engineering fixes, not to switch to larger architectures. Compute is secondary and non-binding: the fine-tune was tractable on modest T4 hardware and would become a constraint only if substantially larger or multimodal architectures were pursued, which the evidence does not warrant. The principal practical constraint is time: transcription and, above all, native-speaker validation take considerably longer than recording. Allocating a dedicated window for collection and validation is therefore essential, as under-resourcing validation tends to yield a larger but noisier dataset. The reusable open-source pipeline released here, comprising the recorder, datasets, fine-tuning recipe, evaluation harness (Senti-Check), and reference application (KasaHealth), establishes that voice-first ASRH delivery in local languages is feasible and well received, de-risks future community-facing investment, and provides a lasting public good for African NLP.

### 4.5 Open-Source Release

All datasets, model weights, and tooling produced in this work are publicly released; Table 5 lists the resources and their locations.

**Table 5. Open-source release inventory.**

| Resource | Link |
|---|---|
| Youth-domain text; Youth ASRH ASR audio datasets | huggingface.co/collections/ghananlpcommunity/{youth-domain-datasets, unicef-asr-datasets} |
| Ghana Bible corpus (~90k) / fine-tuned model weights | huggingface.co/{datasets/ghananlpcommunity/ghana-bible-combined-90k…, ghananlpcommunity/qwen3-asr-0.6b-ghana…} |
| Recorder / Senti-Check / KasaHealth (live) | github.com/GhanaNLP/{recorder, senti-check} · kasa-health-frontend.onrender.com |

## 5. CONCLUSION

This study reports an end-to-end investigation of ASR for adolescent health communication in Twi, Dagbani, and Ewe, spanning benchmarking, domain adaptation, and user validation. The benchmark confirms a clear character-level advantage for language-specific Wav2Vec2 models (CER 17–25%) and identifies Gemma 4 as a competitive zero-shot baseline at the word level, while documenting a persistent domain gap across all models. Following an infeasible Gemma fine-tune, the compact Qwen3-ASR-0.6B was fine-tuned on a large out-of-domain corpus and evaluated on strictly held-out in-domain audio, reducing WER for every language and by 44.5 pp for Ewe. KasaHealth demonstrates that a production local-language system is usable and trusted by community users (100% chat approval, 72% Good-or-Excellent translation, 92% would recommend), and Senti-Check shows that the fine-tune doubles downstream semantic accuracy. The results consistently identify validated in-domain data as the binding constraint. The open-source pipeline released with this work provides a foundation for that investment and for adoption by other African language communities.

**Funding**. The project was funded by UNICEF Office of Innovation (WCARO) with Contract number: 43456330 having duration from 19.March.2026 to 18.June.2026.

**Acknowledgements.** Carried out by the Ghana NLP Community with the UNICEF Office of Innovation (WCARO) and UNICEF Ghana. We thank the community volunteers and native-speaker testers who contributed recordings, transcriptions, and evaluations; Jonathan Asiamah, Issah Abdul Haqq Niendoo, Humphrey Donkor, Emmanuel Adu Saah, Priscilla N. Lartey, Livingstone Eli Ayivor, Datsomor Gerhardt Kwame, Lucas Woedem Kpatah, Mukson Ibrahim, Ibrahim Abdul-Halim, Sualey Naporo Alhassan, Alhassan Amin Naporo, Jida Asare, Franklyn Armah, Agartha Enyonam Aziedor, Benedicta Tutu, Alhassan Fuseini Naporo